\documentclass{ifacconf}

\usepackage{graphicx}      
\usepackage{natbib}        
\usepackage[T1]{fontenc}
\usepackage{textgreek}
\usepackage[noadjust]{cite}

\begin{document}
\begin{frontmatter}
\title{Hydraulic Volumetric Soft Everting Vine Robot Steering
Mechanism for Underwater Exploration}
   
\author[First]{Danyaal Kaleel}
\author[Second]{Benoit Clement}
\author[First]{Kaspar Althoefer}

\address[First]{Centre for Advanced Robotics @ Queen Mary (ARQ), Queen Mary University of London, Mile End Road, London, E1 4NS, United Kingdom (email: {d.m.kaleel, k.althoefer} @qmul.ac.uk)}
\address[Second]{ENSTA Bretagne, IRL CNRS 2010 Crossing, Australia}
© 2024 D. Kaleel, B. Clement, K. Althoefer. This work has been accepted to IFAC for publication under a Creative Commons Licence CC-BY-NC-ND
\maketitle

\begin{abstract}
Despite a significant proportion of the Earth being covered in water, exploration of what lies below has been limited due to the challenges and difficulties inherent in the process. Current state of the art robots such as Remotely Operated Vehicles (ROVs) and Autonomous Underwater Vehicles (AUVs) are bulky, rigid and unable to conform to their environment. This makes certain underwater regions, especially those characterised by tight and narrow crevasses - as is the case for coral reefs - inaccessible to currently available exploratory tools. Soft robotics offers potential solutions to this issue. Fluid-actuated eversion or growing robots, in particular, are a good example. While current eversion robots have found many applications on land, their inherent properties make them particularly well suited to underwater environments. An important factor when considering  underwater eversion robots is the establishment of a suitable steering mechanism that can enable the robot to change direction as required. This project proposes a design for an eversion robot that is capable of steering while underwater, through the use of bending pouches, a design commonly seen in the literature on land-based eversion robots. These bending pouches contract to enable directional change. Similar to their land-based counterparts, the underwater eversion robot uses the same fluid in the medium it operates in to achieve extension and bending but also to additionally aid in neutral buoyancy. The actuation method of bending pouches meant that robots needed to fully extend before steering was possible. Three robots, with the same design and dimensions were constructed from polyethylene tubes and tested. Our research shows that although the soft eversion robot design in this paper was not capable of consistently generating the same amounts of bending for the inflation volume, it still achieved suitable bending at a range of inflation volumes and was observed to bend to a maximum angle of 68 degrees at 2000 ml, which is in line with the bending angles reported for land-based eversion robots in the literature.
\end{abstract}

\begin{keyword}
underwater eversion robot, underwater everting vine robot, underwater eversion robot steering mechanism, underwater soft robot, hydraulic eversion 
\end{keyword}

\end{frontmatter}


\section{Introduction}
Approximately 71\% of the Earth's surface is covered in water. Despite this huge expanse of area, exploration under the surface of these bodies of water has been limited due to humans' inability to spend prolonged periods underwater. While scuba equipment facilitates exploration of shallow waters where pressure values are lower, oxygen supply remains a critical factor, in turn limiting time spent underwater. Greater depths require specialist equipment and are unsuitable environments for humans due to the immense pressure. Manned submersibles are an option but these too have limited operational time, and due to their physical bulk cannot manoeuvre in complex environments. An autonomous solution that mitigates human risk yet enables intricate navigation is therefore required.

Robotic systems can provide a route to solving this problem - indeed they are already being used in this way. Known as Remotely Operated Vehicles (ROVs), they are essentially tethered underwater robots, \citep{petillot2019underwater}, and have been used for a variety of applications including exploration \citep{macreadie2018eyes},  biological \citep{chaloux2021novel} and geological \citep{paull2001deep} sample collection, and marine renewable installation and inspection \citep{elvander2012rovs}. Despite these robots being widely used across the marine industry, they have a number of disadvantageous properties that render them unsuitable for operating in underwater environments that are small, narrow, tight and fragile. These robots can weigh a considerable amount, between 100 kg and 5000 kg for intervention type ROVs and between 3 kg and 120 kg for inspection ROVs \citep{capocci2017inspection}. Although they range in size, most of them tend to be large and constructed from non-environmentally compliant, rigid materials. This rigidity makes it difficult for the robot to adapt to its environment and its large size also limits the underwater environments within which it can be used. These robots tend to also be expensive to develop and produce, especially if required to go to depths of over 100 metres \citep{teague2018potential}. Operational costs of ROVs also tend to be high, as they need a crewed support ship to which they are tethered during use \citep{schjolberg2015towards}. Soft robots are naturally adaptable to their environment due to their constituent deformable soft materials. This can help in minimising the amount of control that is needed to mitigate damage to fragile marine ecosystems such as coral reefs. Soft robots are also relatively cheap to manufacture, on account of their basic materials. Their loss, therefore, would not be catastrophic.
One type of soft robot known as an eversion robot or everting vine robot, as seen in \citep{hawkes2017soft}, is cylindrical in shape, typically constructed from fabrics \citep{putzu2018plant} or thin plastic sheets \citep{blumenschein2018tip}, that moves through its environment by way of fluid-actuated extensions of its body. These robots can carry tools such as cameras, sensors or payload collection mechanisms to a target point in their environment using cap mechanisms such as \citep{suulker2023soft} that sit at the tip of the robot, while continually moving forward.
There are a considerable number of potential applications that could benefit from such a robot.
These include coral reef exploration, ship, offshore wind turbine and oil rig inspection, environmental monitoring, mine clearance \citep{kaleel2022underwater} and for archaeological discovery work such as on ship wrecks.
Environments that are cluttered with obstacles are not a problem for this type of robot either as it is able to navigate around them with ease, as seen in \citep{greer2020robust}.

At the time of writing, there are a few examples in the literature of eversion robots in underwater environments, among them (\citep{tennakoon2023experimental} and \citep{luong2019eversion}). \citep{luong2019eversion} developed an eversion robot that used hydraulic actuation to extend and retract underwater and change direction using environmental contact. \citep{tennakoon2023experimental} looked into characterising the amount of force the eversion robot can exert underwater and the velocity it can achieve, during extension, at different depths. Neither paper showed an eversion robot able to choose the direction in which it steers underwater. Having a method to actively control the direction in which the robot moves is important for real-world applications.

\begin{figure}
\centering
\includegraphics[width = \linewidth]{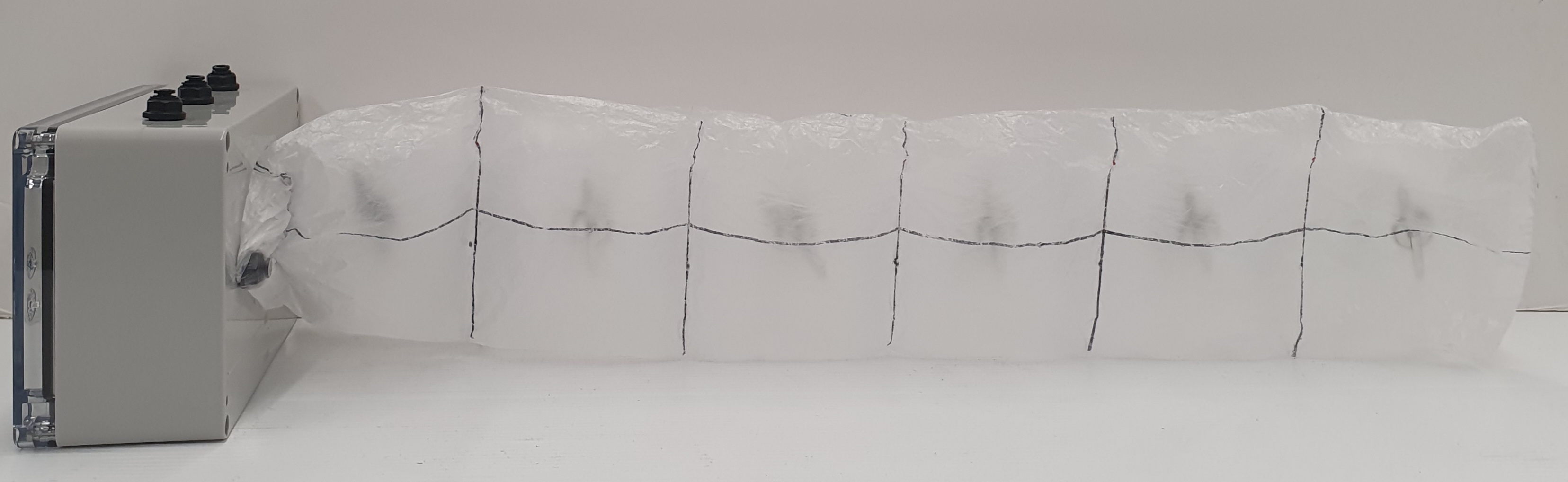}
  \caption{Side view of robot}
  \label{fig 9}
\end{figure}

The work shown in this paper aims to develop an eversion robot system that is capable of steering left and right underwater (Fig. \ref{fig 9}). Unlike the numerous pneumatically operated eversion robot steering mechanisms seen in the literature for land-based eversion robots, which use air pressure to cause bending, water does not achieve bending through pressure change but through volume change. As far as the authors of this paper are aware, this is the first ever volume pouch based steering mechanism for an eversion robot and indeed the first time that any kind of steering mechanism has been developed for an underwater eversion robot. Therefore, the scientific question that this paper answers is how volumetric inflation affects the bending angle of the robot. The robot in this paper had its main everting chamber filled with water, no different in composition to the surrounding water, making it neutrally buoyant. It had bending pouches located on the left and right of its main everting chamber, which enabled the robot to turn as they contracted when filled with water. 

The paper is structured as follows: Section 2 gives an overview of typical eversion robot steering mechanisms found in the literature, how these mechanisms differ from one another, and why our chosen method is suitable for the kind of applications we are looking to achieve. Section 3 looks into the design of the robot - notably its dimensions, material properties and construction. Section 4 discusses our experimental setup, including the testing apparatus used, how the experiment was carried out and how the data was analysed. Section 5 describes the results focusing on the range of bending angles that are achievable for different volumes of water and the maximum bending angle. Section 6 draws conclusions and makes recommendations relating to the direction of future research.

\section{Background of steering mechanisms}
There are three categories of steering mechanism for eversion robots; tendon driven steering such as \citep{gan20203d}, inner skeleton steering such as \citep{wu2023vision} and inflated pouch steering such as \citep{ataka2020model}.

Tendon driven steering uses wires that pull along the length of the robot body from base to tip. They are analogous to the reins for a horse and dictate the robot's direction. This approach is also used in rigid-body robots; examples include the Barrett hand \citep{townsend2000barretthand} and the da Vinci Surgical System \citep{douissard2019vinci}. Eversion robots of this type are unable to produce more than one or two bends limiting the number of movements that can be achieved.

Inner skeleton steering uses a mechanism that is placed inside, but not attached to the everting body material \citep{takahashi2022inflated}. This skeleton is then actuated via one of four main methods, these being: hinge-based motor driven \citep{haggerty2021hybrid}, hinge-based tendon driven \citep{takahashi2021eversion}, tendon-driven catheter skeleton \citep{berthet2021mammobot} and actuated backbone \citep{der2021roboa}.  
The hinge-based mechanisms (\citep{haggerty2021hybrid} and \citep{takahashi2021eversion}) consist of two rigid tubes connected with a hinge. By changing the orientation of the hinge, the direction of steering can be set. The catheter mechanisms (\citep{berthet2021mammobot} and \citep{wu2023vision}) steer the robot by bending at the tip of the robot such that the robot body follows its bending angle.
The actuated backbone (\citep{der2021roboa}) is a rigid structure that bends itself and in doing so bends the everting material around it.
A limitation of the two hinge based mechanisms and the actuated backbone mechanism is that they contain rigid components. This could present problems if the robot needed to navigate through a constriction, which has a smaller diameter than the rigid body parts. An underwater example might be narrowings in coral reefs. A limitation of the catheter and tendon driven mechanisms is that the robot can only be steered up to the length of the catheter or tendon. This means that once the robot has grown past the length of the catheter or tendon, steering is no longer controllable. A further problem with all these inner skeleton mechanisms is that they do not hold their shape once the skeleton moves forward - potentially a  problem in applications where bending needs to be maintained for the duration of operation.

An inflated pouch steering mechanism is one that is attached to the side of the robot's body, either integrated into the walls \citep{abrar2021highly} or attached externally \citep{greer2019soft}. When the pouches are inflated, they contract outward, causing the outer edge to become smaller than the inner edge. This in turn makes the robot bend towards the contraction. Robots that make use of this system are able to maintain their shape as the robot extends forward and are able to produce multiple bends, dependent upon how many pouches or sets of pouches the robot has. As they have no constituent rigid components, navigation through tight spaces becomes possible. 

Given the advantageous properties highlighted above, the inflated pouch steering mechanism was chosen for further investigation in relation to underwater applications.  

\section{Design and Build}
\subsection{Robot Anatomy}
\begin{figure}
\centering
\includegraphics[width = \linewidth]{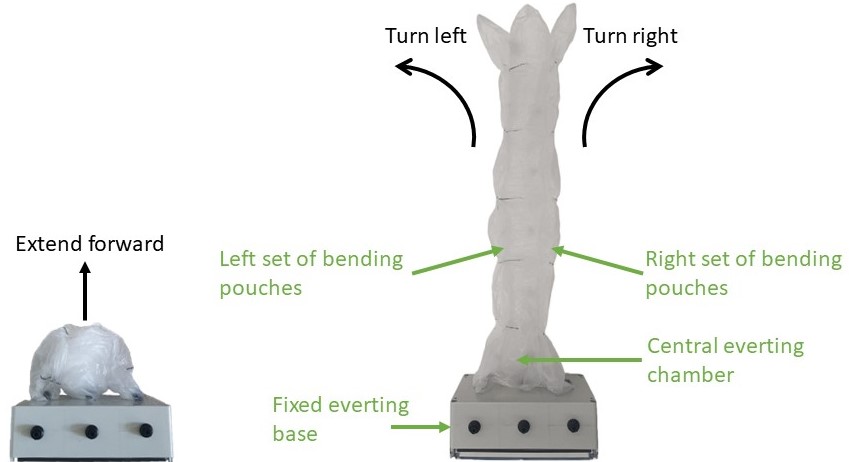}
  \caption{(Left) Top view of robot in retracted state showing the extension movement that it is capable of producing in this state. (Right) Top view of robot in everted state showing the bending movements that it is capable of (black arrows and text) and the 4 main parts that make up the robot (green arrows and text).}
  \label{fig 6}
\end{figure}

The robot consists of a fixed base (Fig. \ref{fig 6}), from which the robot extends, using a construction similar to that of \citep{kaleel2023framework}, and three flexible polyethylene tubes (Cabilock Transparent Disposable Umbrella Rain Bags) as seen in Fig. \ref{fig 6}. Referring to this figure, the middle polyethylene tube is for extension or eversion, and both the left and right polyethylene tubes are manufactured into bending pouches which have partial constrictions along their length to enable steering to the left or right. The robot weighs 368 grams when not inflated with water.

\subsection{Manufacturing Process}
\subsubsection{Fixed Robot Base}
\begin{figure}[h]
\centering
\includegraphics[width = \linewidth]{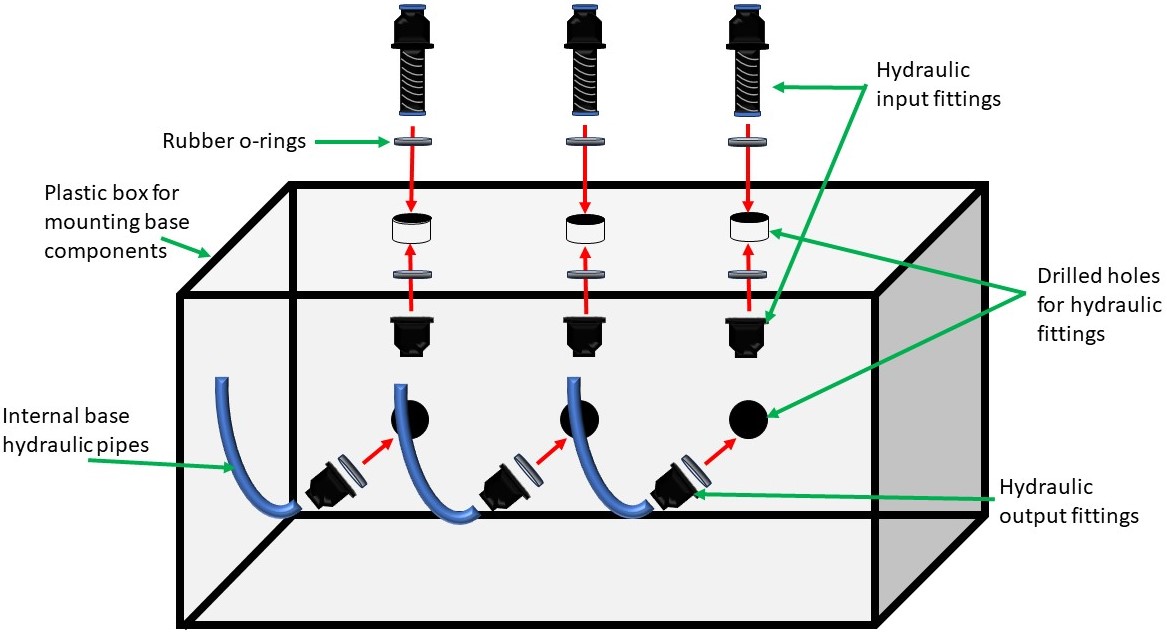}
  \caption{Construction of the fixed robot base}
  \label{fig 13}
\end{figure}

The robot's fixed base consisted of a plastic box (Fibox Polycarbonate Enclosure), hydraulic pipe fittings (RS PRO Push-in Fitting 4mm to 4mm), hydraulic pipes (SMC Compressed Air Pipe 4mm Outer Diameter) and o-rings (RS PRO Nitrile Rubber O-ring, 10mm Bore, 12 mm Outer Diameter), as seen in Fig. \ref{fig 13}.
Six holes were drilled to accommodate the pipe fittings. O-rings were then placed between the fittings, to create a tight seal, and the fittings then screwed into these holes. Three of the holes were drilled on the top of the box for hydraulic input, and connected to a water supply. The three other holes were drilled on the front surface of the box for hydraulic output to the soft flexible part of the robot. These fittings were also used to connect the soft flexible material of the robot to the fixed base. Internal hydraulic pipes within the base connected the input fittings to the output fittings.

\subsubsection{Assembly of Soft Flexible Robot}
\begin{figure*}
\centering
\includegraphics[width = 0.75\linewidth]{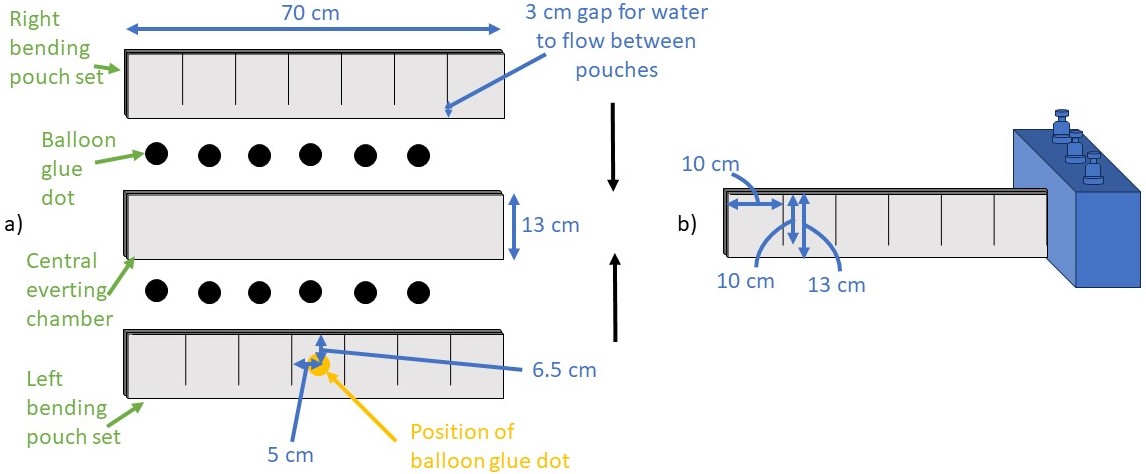}
  \caption{(a) The assembly of the everting section of the robot from three polyethylene tubes and the position of the balloon glue in relation to the bending pouches. (b) This figure shows the assembled eversion robot with its base and the dimensions of the bending pouches.}
  \label{fig 4}
\end{figure*}

The soft part of the robot was formed from three 70 cm long plastic tubes, each with a circumference of 26 cm (see Fig. \ref{fig 4}). 

The first step in the manufacturing process was to mark out, using a pen, the sizes and positions of the pouches on two of the polyethylene tubes that would form the two sets of volumetric bending pouches. One set of pouches was for bending horizontally left and the other for bending horizontally right. A plastic bag sealer was then used to create partial constrictions along the length of the bending pouches through plastic welding. Partial constrictions were made so that water could flow between individual pouches in a pouch set, enabling the pouches to expand and contract, resulting in bending. Each pouch was uniform in size and square in shape - 10 cm long and 10 cm wide with a 3 cm gap for water to flow between the individual pouches in a set (Fig. \ref{fig 4}). This created a set of seven individual pouches for each bending pouch set. Note that the central everting tube was left in its cylindrical state to enable extension. 

After creating the two sets of bending pouches, the everting section of the robot was assembled. An illustration of the steps required in assembling the individual components of the everting section of the robot is shown in Fig. \ref{fig 4} a).  Note that before assembly, it is advisable to test the bending pouch sets with compressed air at low pressure to ensure they are tightly sealed and bend as expected. Balloon glue (GYTFOG Balloon Glue Dots) were used to connect the middle everting tube to the two sets of bending pouches at discrete points, shown in Fig. \ref{fig 4} a). 

\subsubsection{Connecting the Fixed Base to the Soft Flexible Robot}
The final stage in the build process involved pushing some of the soft material of each of the three tubes through the drilled hole in the base and holding it in place with the output pipe fittings. This meant that for the two bending pouch sets, which had seven pouches, one pouch was used to fix it to the base. This meant that only six of the pouches were used for actuation. 

\section{Experimental Setup}

\subsection{Test Procedure}
\begin{figure}
\centering
\includegraphics[width = 0.95\linewidth]{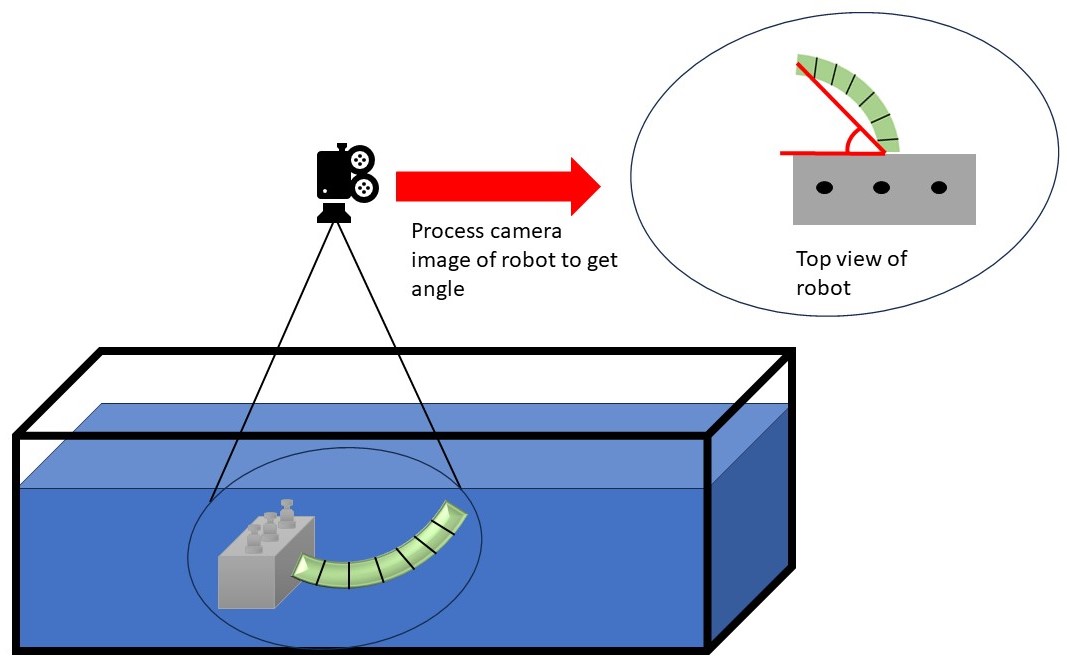}
  \caption{Experimental setup to obtain bending angles based on volume of water.}
  \label{fig 11}
\end{figure}

A tank was filled with water and the robot was placed in the tank in an extended state. One of the two sets of bending pouches was then chosen to be used to carry out the test. Given their identical either one could be used. Then the pouch was filled with 400 ml at each iteration of the test up to a maximum of 3200 ml. An RS M400 micro pump was used to pump water from the input source to the robot. It was powered by a voltage between 5-6V. The rate of the pump determines the speed with which the robot bends and changes shape. To observe the changes in bending angle, a camera suspended above the tank, as shown in Fig. \ref{fig 11}, was used to capture images of the robot, at each 400 ml iteration from 0 to 3200 ml. 

\subsection{Bending Angle Analysis}
The bending angle was measured by drawing two lines; a horizontal line at the edge of the base as a reference line and another to connect the centre of the robot to the tip of the robot as shown in Fig. \ref{fig 11}. A protractor was then used to measure the bending angle between these two lines. The first image taken, which is the image where the robot is not inflated and has 0 ml is used as the reference image to observe and calculate the bending angle from.

\section{Results}
\begin{figure}
\centering
\includegraphics[width = 0.50\linewidth, height = 0.50\linewidth]{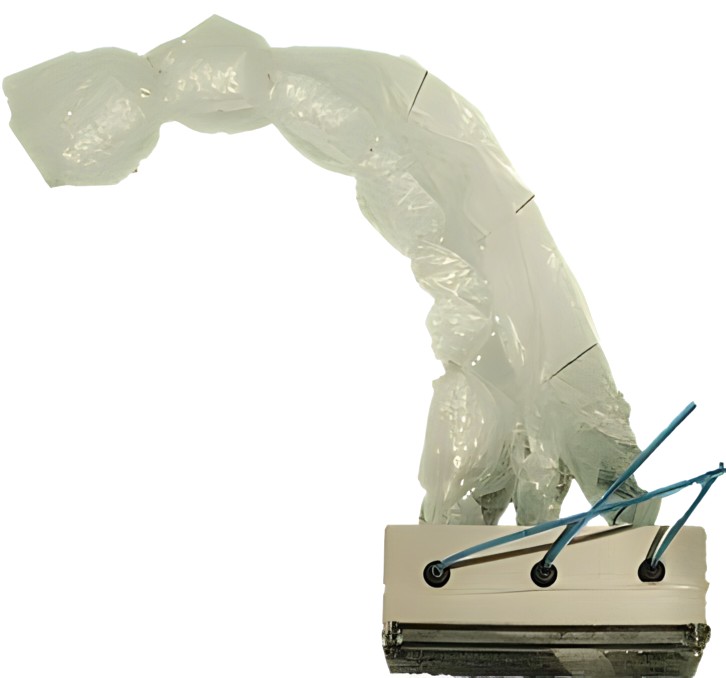}
  \caption{Robot demonstrating an angle of 65 degrees when inflated to the maximum volume of 3228 ml.}
  \label{fig 3}
\end{figure}

The two factors we wanted to determine were (1) the maximum bending angle that is achievable and with what volume of water it is achieved and (2) how the bending angle varies with the volume of water. Note that the middle eversion chamber was extended but did not have any water in it to observe the effect of bending without additional stiffness. Furthermore, as the two sides of the robot are identical, only one set of bending pouches needed testing. The non-tested side could be assumed to mirror the behaviour of the tested side. 
Three different robot prototypes constructed from the same materials and with the same dimensions were used to perform the tests. An example of the robot bending is shown in Fig. \ref{fig 3}.

\subsection{Maximum bending angle for volumetric pouches}
The maximum bending angle achieved by this robot design using volumetric pouch inflation was 68 degrees (Fig. \ref{fig 12}). This was achieved with an inflation volume of 2000 ml, which robot 3 demonstrated. However, from fig. \ref{fig 12}, we can see that the maximum bending angle is not an accurate measurement of the robot's design due to the great variability in each robot's maximum bending angle and at what volume that maximum bending angle was achieved.

\subsection{How bending angle varies with volume}
\begin{figure}[h]
\centering
\includegraphics[width = \linewidth]{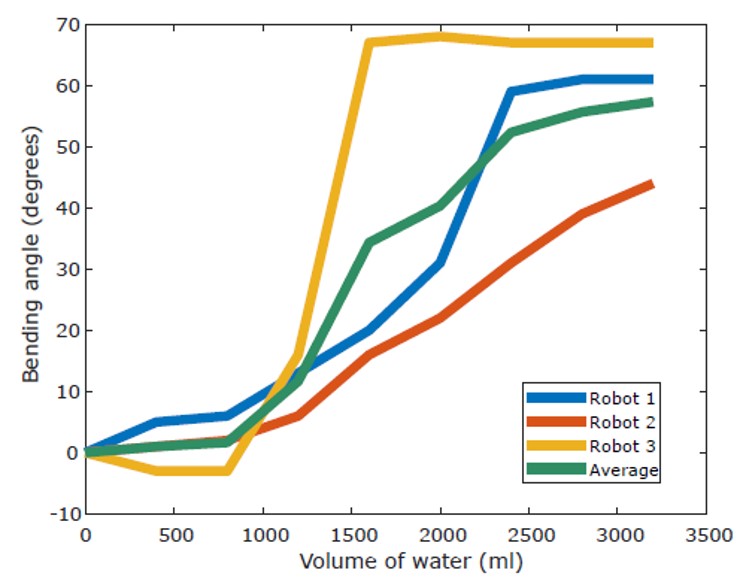}
  \caption{How bending angle varies with volume.}
  \label{fig 12}
\end{figure}

Fig. \ref{fig 12} shows how the bending angle varied with the volume of water used. From this we can see that the bending angle varied greatly between different robot prototypes. The only conclusive similarity between prototypes that we can draw is that an increase in bending angle occurred roughly between 800 ml and plateaued around 2400 ml, although robot 3 plateaued at around 1600 ml. This suggests that varying the amount of water within this range has a greater impact on the robot's bending angle compared to volume changes outside of this range. 

The fact that an increase in bending angle as inflation volume increases is inconsistent between robot's is to be expected due to the non-linear nature of soft robots. Several factors could account for this including vibrations in the environment, differences in pump flow rates, water temperature and varying amounts of air in the robot. Furthermore, each test had a different robot prototype due to the material developing small holes after each experiment. This may suggest that small manufacturing differences between the robots could have accounted for the deviation observed. Despite these results, this experiment still demonstrates the robot's ability to bend to a suitable degree for it to be able to usefully move through its environment. 

\section{Conclusions and Future Work}
The work in this paper set out to develop a volumetric pouch steering mechanism for an underwater eversion robot. It was evaluated against two criteria: (1) the maximum bending angle and (2) how bending angle varied with volume. Three different robot prototypes all with the same designs and dimensions were used to determine these criteria. 
The maximum bending angle observed was 68 degrees at an inflation volume of 2000 ml. This is a good amount of bending when compared to other eversion robot designs that use pouch inflation.
The robot demonstrated it's ability to achieve angles between the initial position and the maximum bending angle. 
Although the results were fairly coherent in that the robots' bending angles increased slowly from 0 ml to 800 ml, increased rapidly as the inflation volume increased upto 2400 ml and then plateaued, the specific values of bending at each volume are different for each robot and cannot be estimated for a specific robot prototype. This is due to the highly non-linear and unpredictable nature of soft eversion robots. This is an important consideration as the robots were tested in an environment that had no controlled variables such as water temperature, which enabled us to observe how it would bend if deployed outside of a controlled setting where 'perfect' conditions do not exist. Despite soft robot motion control being quite challenging to generalise across prototypes with the same properties and design, they still have plenty of desirable properties as discussed in the introduction that make them well worth pursuing and developing further. 

Future work should look into developing a feedback control system for this robotic system. This controller would need to be adaptive to the robot and may require a calibration procedure before the robot can carry out its task. A control system that makes use of reinforcement learning could also be an interesting approach to explore. Future work is also needed to address the manufacturing process and construction of the robot. Specifically, in regards to reducing leakage through small holes in the plastic sheet material caused by the heat welding process, which affected the volume of water in the pouches and the bending angles achievable. One could look to the work presented in \citep{berthet2021mammobot} for inspiration on a heat welding mechanism that could prevent manufacturing inconsistencies and flaws. One could also look into using an alternative material, which is still flexible enough for eversion but more resistant to higher pressure. Thirdly work is needed to characterise the dynamic response of the system and it's ability to bend under different water conditions such as salty water, water of different temperatures and water at different pressures. Finally, the robot in our paper had air retained within it, which meant that it was not fully submerged. One could look into improving the design such as adding suitable weights along the length of the robot to counteract the upward force that the air generates or alternatively look at a method of removing the air entirely.

\section{Acknowledgements}
This work is funded by a joint program between French AID (Agence Innovation Defence) and British DSTL. Thank you to our collaborator Dr Adrian Baker from DSTL for his help on the work presented and to Mish Toszeghi for his help in the preparation of this paper.

\bibliography{ref}            

\begin{thebibliography}{29}
\providecommand{\natexlab}[1]{#1}
\providecommand{\url}[1]{\texttt{#1}}
\providecommand{\urlprefix}{URL }
\expandafter\ifx\csname urlstyle\endcsname\relax
  \providecommand{\doi}[1]{doi:\discretionary{}{}{}#1}\else
  \providecommand{\doi}{doi:\discretionary{}{}{}\begingroup \urlstyle{rm}\Url}\fi

\bibitem[{Abrar et~al.(2021)Abrar, Putzu, Ataka, Godaba, and Althoefer}]{abrar2021highly}
Abrar, T., Putzu, F., Ataka, A., Godaba, H., and Althoefer, K. (2021).
\newblock Highly manoeuvrable eversion robot based on fusion of function with structure.
\newblock In \emph{2021 IEEE International Conference on Robotics and Automation (ICRA)}, 12089--12096. IEEE.

\bibitem[{Ataka et~al.(2020)Ataka, Abrar, Putzu, Godaba, and Althoefer}]{ataka2020model}
Ataka, A., Abrar, T., Putzu, F., Godaba, H., and Althoefer, K. (2020).
\newblock Model-based pose control of inflatable eversion robot with variable stiffness.
\newblock \emph{IEEE Robotics and Automation Letters}, 5(2), 3398--3405.

\bibitem[{Berthet-Rayne et~al.(2021)Berthet-Rayne, Sadati, Petrou, Patel, Giannarou, Leff, and Bergeles}]{berthet2021mammobot}
Berthet-Rayne, P., Sadati, S.H., Petrou, G., Patel, N., Giannarou, S., Leff, D.R., and Bergeles, C. (2021).
\newblock Mammobot: A miniature steerable soft growing robot for early breast cancer detection.
\newblock \emph{IEEE Robotics and Automation Letters}, 6(3), 5056--5063.

\bibitem[{Blumenschein et~al.(2018)Blumenschein, Gan, Fan, Okamura, and Hawkes}]{blumenschein2018tip}
Blumenschein, L.H., Gan, L.T., Fan, J.A., Okamura, A.M., and Hawkes, E.W. (2018).
\newblock A tip-extending soft robot enables reconfigurable and deployable antennas.
\newblock \emph{IEEE Robotics and Automation Letters}, 3(2), 949--956.

\bibitem[{Capocci et~al.(2017)Capocci, Dooly, Omerdi{\'c}, Coleman, Newe, and Toal}]{capocci2017inspection}
Capocci, R., Dooly, G., Omerdi{\'c}, E., Coleman, J., Newe, T., and Toal, D. (2017).
\newblock Inspection-class remotely operated vehicles—a review.
\newblock \emph{Journal of Marine Science and Engineering}, 5(1), 13.

\bibitem[{Chaloux et~al.(2021)Chaloux, Phillips, Gruber, Schelly, and Sparks}]{chaloux2021novel}
Chaloux, N., Phillips, B.T., Gruber, D.F., Schelly, R.C., and Sparks, J.S. (2021).
\newblock A novel fish sampling system for rovs.
\newblock \emph{Deep Sea Research Part I: Oceanographic Research Papers}, 167, 103428.

\bibitem[{Der~Maur et~al.(2021)Der~Maur, Djambazi, Haberth{\"u}r, H{\"o}rmann, K{\"u}bler, Lustenberger, Sigrist, Vigen, F{\"o}rster, Achermann, Hampp, Katzschmann, and Siegwart}]{der2021roboa}
Der~Maur, P.A., Djambazi, B., Haberth{\"u}r, Y., H{\"o}rmann, P., K{\"u}bler, A., Lustenberger, M., Sigrist, S., Vigen, O., F{\"o}rster, J., Achermann, F., Hampp, E., Katzschmann, R.K., and Siegwart, R. (2021).
\newblock Roboa: Construction and evaluation of a steerable vine robot for search and rescue applications.
\newblock In \emph{IEEE 4th International Conference on Soft Robotics (RoboSoft)}, 15--20. IEEE.

\bibitem[{Douissard et~al.(2019)Douissard, Hagen, and Morel}]{douissard2019vinci}
Douissard, J., Hagen, M.E., and Morel, P. (2019).
\newblock The da vinci surgical system.
\newblock \emph{Bariatric robotic surgery: a comprehensive guide}, 13--27.

\bibitem[{Elvander and Hawkes(2012)}]{elvander2012rovs}
Elvander, J. and Hawkes, G. (2012).
\newblock Rovs and auvs in support of marine renewable technologies.
\newblock In \emph{2012 Oceans}, 1--6. IEEE.

\bibitem[{Gan et~al.(2020)Gan, Blumenschein, Huang, Okamura, Hawkes, and Fan}]{gan20203d}
Gan, L.T., Blumenschein, L.H., Huang, Z., Okamura, A.M., Hawkes, E.W., and Fan, J.A. (2020).
\newblock 3d electromagnetic reconfiguration enabled by soft continuum robots.
\newblock \emph{IEEE Robotics and Automation Letters}, 5(2), 1704--1711.

\bibitem[{Greer et~al.(2020)Greer, Blumenschein, Alterovitz, Hawkes, and Okamura}]{greer2020robust}
Greer, J.D., Blumenschein, L.H., Alterovitz, R., Hawkes, E.W., and Okamura, A.M. (2020).
\newblock Robust navigation of a soft growing robot by exploiting contact with the environment.
\newblock \emph{The International Journal of Robotics Research}, 39(14), 1724--1738.

\bibitem[{Greer et~al.(2019)Greer, Morimoto, Okamura, and Hawkes}]{greer2019soft}
Greer, J.D., Morimoto, T.K., Okamura, A.M., and Hawkes, E.W. (2019).
\newblock A soft, steerable continuum robot that grows via tip extension.
\newblock \emph{Soft robotics}, 6(1), 95--108.

\bibitem[{Haggerty et~al.(2021)Haggerty, Naclerio, and Hawkes}]{haggerty2021hybrid}
Haggerty, D.A., Naclerio, N.D., and Hawkes, E.W. (2021).
\newblock Hybrid vine robot with internal steering-reeling mechanism enhances system-level capabilities.
\newblock \emph{IEEE Robotics and Automation Letters}, 6(3), 5437--5444.

\bibitem[{Hawkes et~al.(2017)Hawkes, Blumenschein, Greer, and Okamura}]{hawkes2017soft}
Hawkes, E.W., Blumenschein, L.H., Greer, J.D., and Okamura, A.M. (2017).
\newblock A soft robot that navigates its environment through growth.
\newblock \emph{Science Robotics}, 2(8), eaan3028.

\bibitem[{Kaleel et~al.(2022)Kaleel, Clement, and Althoefer}]{kaleel2022underwater}
Kaleel, D., Clement, B., and Althoefer, K. (2022).
\newblock Underwater eversion robot growth for underwater operations with an emphasis on underwater mine hunting.
\newblock In \emph{IEEE UK\&I RAS Conference}.

\bibitem[{Kaleel et~al.(2023)Kaleel, Clement, and Althoefer}]{kaleel2023framework}
Kaleel, D., Clement, B., and Althoefer, K. (2023).
\newblock A framework to design and build a height controllable eversion robot.
\newblock In \emph{2023 11th International Conference on Control, Mechatronics and Automation (ICCMA)}, 239--244. IEEE.

\bibitem[{Luong et~al.(2019)Luong, Glick, Ong, DeVries, Sandin, Hawkes, and Tolley}]{luong2019eversion}
Luong, J., Glick, P., Ong, A., DeVries, M.S., Sandin, S., Hawkes, E.W., and Tolley, M.T. (2019).
\newblock Eversion and retraction of a soft robot towards the exploration of coral reefs.
\newblock In \emph{2019 2nd IEEE International Conference on Soft Robotics (RoboSoft)}, 801--807. IEEE.

\bibitem[{Macreadie et~al.(2018)Macreadie, McLean, Thomson, Partridge, Jones, Gates, Benfield, Collin, Booth, Smith et~al.}]{macreadie2018eyes}
Macreadie, P.I., McLean, D.L., Thomson, P.G., Partridge, J.C., Jones, D.O., Gates, A.R., Benfield, M.C., Collin, S.P., Booth, D.J., Smith, L.L., et~al. (2018).
\newblock Eyes in the sea: unlocking the mysteries of the ocean using industrial, remotely operated vehicles (rovs).
\newblock \emph{Science of the Total Environment}, 634, 1077--1091.

\bibitem[{Paull et~al.(2001)Paull, Stratton, Conway, Brekke, Dawe, Maher, and Ussler}]{paull2001deep}
Paull, C., Stratton, S., Conway, M., Brekke, K., Dawe, T.C., Maher, N., and Ussler, W. (2001).
\newblock Deep sea vibracoring system improves rov sampling capability.
\newblock \emph{Eos, Transactions American Geophysical Union}, 82(30), 325--326.

\bibitem[{Petillot et~al.(2019)Petillot, Antonelli, Casalino, and Ferreira}]{petillot2019underwater}
Petillot, Y.R., Antonelli, G., Casalino, G., and Ferreira, F. (2019).
\newblock Underwater robots: From remotely operated vehicles to intervention-autonomous underwater vehicles.
\newblock \emph{IEEE Robotics \& Automation Magazine}, 26(2), 94--101.

\bibitem[{Putzu et~al.(2018)Putzu, Abrar, and Althoefer}]{putzu2018plant}
Putzu, F., Abrar, T., and Althoefer, K. (2018).
\newblock Plant-inspired soft pneumatic eversion robot.
\newblock In \emph{2018 7th IEEE International Conference on Biomedical Robotics and Biomechatronics (Biorob)}, 1327--1332. IEEE.

\bibitem[{Schj{\o}lberg and Utne(2015)}]{schjolberg2015towards}
Schj{\o}lberg, I. and Utne, I.B. (2015).
\newblock Towards autonomy in rov operations.
\newblock \emph{IFAC-PapersOnLine}, 48(2), 183--188.

\bibitem[{Suulker et~al.(2023)Suulker, Skach, Kaleel, Abrar, Murtaza, Suulker, and Althoefer}]{suulker2023soft}
Suulker, C., Skach, S., Kaleel, D., Abrar, T., Murtaza, Z., Suulker, D., and Althoefer, K. (2023).
\newblock Soft cap for vine robots.
\newblock In \emph{2023 IEEE/RSJ International Conference on Intelligent Robots and Systems (IROS)}, 6462--6468. IEEE.

\bibitem[{Takahashi et~al.(2021)Takahashi, Tadakuma, Watanabe, Takane, Hookabe, Kajiahara, Yamasaki, Konyo, and Tadokoro}]{takahashi2021eversion}
Takahashi, T., Tadakuma, K., Watanabe, M., Takane, E., Hookabe, N., Kajiahara, H., Yamasaki, T., Konyo, M., and Tadokoro, S. (2021).
\newblock Eversion robotic mechanism with hydraulic skeletonto realize steering function.
\newblock \emph{IEEE Robotics and Automation Letters}, 6(3), 5413--5420.

\bibitem[{Takahashi et~al.(2022)Takahashi, Watanabe, Abe, Tadakuma, Saiki, Konyo, and Tadokoro}]{takahashi2022inflated}
Takahashi, T., Watanabe, M., Abe, K., Tadakuma, K., Saiki, N., Konyo, M., and Tadokoro, S. (2022).
\newblock Inflated bendable eversion cantilever mechanism with inner skeleton for increased stiffness.
\newblock \emph{IEEE Robotics and Automation Letters}, 8(1), 168--175.

\bibitem[{Teague et~al.(2018)Teague, Allen, and Scott}]{teague2018potential}
Teague, J., Allen, M.J., and Scott, T.B. (2018).
\newblock The potential of low-cost rov for use in deep-sea mineral, ore prospecting and monitoring.
\newblock \emph{Ocean Engineering}, 147, 333--339.

\bibitem[{Tennakoon et~al.(2023)Tennakoon, Subasingha, Sewwandhi, Kulasekera, and Dassanayake}]{tennakoon2023experimental}
Tennakoon, K.E., Subasingha, G.S., Sewwandhi, J.A., Kulasekera, A.L., and Dassanayake, P.C. (2023).
\newblock Experimental performance characterization of an underwater growing robot.
\newblock In \emph{2023 Moratuwa Engineering Research Conference (MERCon)}, 620--625. IEEE.

\bibitem[{Townsend(2000)}]{townsend2000barretthand}
Townsend, W. (2000).
\newblock The barretthand grasper--programmably flexible part handling and assembly.
\newblock \emph{Industrial Robot: an international journal}, 27(3), 181--188.

\bibitem[{Wu et~al.(2023)Wu, Sadati, Rhode, and Bergeles}]{wu2023vision}
Wu, Z., Sadati, S.H., Rhode, K., and Bergeles, C. (2023).
\newblock Vision-based autonomous steering of a miniature eversion growing robot.
\newblock \emph{IEEE Robotics and Automation Letters}.

\end{thebibliography}

\end{document}